\documentclass{article}
\usepackage{float}
\usepackage[top=1.25in, bottom=1.25in, left=1.25in, right=1.25in]{geometry}
\usepackage{booktabs}
\usepackage{graphicx}
\usepackage{url}
\usepackage{amsmath}
\usepackage{amssymb}

\title{How Far is Adam from Natural Gradient Descent?}
\author{\textbf{Vihaan Paka-Hegde}$^1$ \\
\texttt{vihaan.paka-hegde\_29@sfuhs.org}}
\date{San Francisco University High School$^1$}

\begin{document}
\maketitle

\begin{abstract}
Adam is the standard optimizer in deep learning, yet its geometric relationship to natural gradient descent (NGD) contains unresolved questions. We study Adam's full update rule, including momentum, as a diagonal empirical Fisher approximation subject to diagonal truncation, empirical label substitution, and temporal lag. Using the scale-invariant \(\gamma(\Delta\theta)\) metric, we measure Adam’s geometric deviation from true NGD across four loss landscapes: well-conditioned linear regression, ill-conditioned linear regression, logistic regression, and a non-convex small neural network. Adam’s geometric trajectory is context-dependent. Deviation remains low in well-conditioned settings but rises significantly under ill-conditioning, reaching misalignments of \(\approx 10^3\) in the neural network. Higher geometric drift correlates with slower initial optimization but does not degrade final objective minimization; Adam consistently reaches low loss. Furthermore, the improved empirical Fisher (iEF) tracks more stable paths than the standard empirical Fisher (EF), which frequently oscillates or diverges. Our results suggest Adam's practical optimization power may stem from a balance of structural approximation errors and momentum smoothing rather than close tracking of the natural gradient path.
\end{abstract}

\section{Introduction}
Modern deep learning relies on Adam due to its robust empirical performance across complex loss landscapes \cite{kingma2014}. Kingma \& Ba motivated this formulation by noting that its coordinate-wise preconditioner approximates the diagonal of the exact Fisher information matrix \cite{kingma2014}. However, treating Adam as a geometric relative of natural gradient descent (NGD) remains contested. Kunstner et al. showed that the empirical Fisher can distort the true gradient field far from the optimum \cite{kunstner2019}, and Wu et al. isolated the inversely-scaled projection issue as a critical failure mode in empirical approximations \cite{wu2024}. 

While Wu et al. introduced an improved empirical Fisher (iEF) to correct full-matrix pathologies \cite{wu2024}, the optimization effects of these errors within Adam's diagonal format remain unisolated. Hwang proposed an analytical bridge through the FAdam variant \cite{hwang2024}, but that framework omits momentum by setting $\beta_1 = 0$. We apply Wu et al.'s geometric framework \cite{wu2024} directly to Adam's complete, unsimplified update rule to measure where diagonal approximations diverge from true NGD across varied landscapes, testing whether iEF corrections preserve stability when coupled with active momentum.

\section{Background}

\subsection{Natural Gradient Descent}
Standard gradient descent treats the parameter space as a flat Euclidean vector space, making optimization sensitive to arbitrary parameterizations. To address this, Amari constructed natural gradient descent (NGD), which models the parameter space as a Riemannian manifold based on the information geometry of the model's output distributions \cite{amari1998}. Given a parametric model family $p(y|x,\theta)$ and a data distribution $p_{\text{data}}(x)$, NGD premultiplies the Euclidean gradient $\nabla_\theta \mathcal{L}(\theta)$ by the inverse of the exact Fisher information matrix $\mathbf{F}$ \cite{amari1998}, defined as

\begin{equation}
\mathbf{F} = \mathbb{E}_{x \sim p_{\text{data}}} 
\mathbb{E}_{y \sim p(y|x,\theta)} \left[
\nabla_\theta \log p(y|x,\theta)\,
\nabla_\theta \log p(y|x,\theta)^\top
\right].
\end{equation}

Computing exact NGD has a high computational cost for high-dimensional models, requiring $\mathcal{O}(D^2)$ space for storage and $\mathcal{O}(D^3)$ time for inversion at each step.

\subsection{The Empirical Fisher Approximation}
Practical natural gradient variants use two primary approximations. First, they substitute the expectation over model-generated predictions with empirical data observations \cite{kunstner2019}, yielding the empirical Fisher (EF) matrix

\begin{equation}
\hat{\mathbf{F}}^{\text{EF}} = \frac{1}{N} \sum_{i=1}^{N}
\nabla_\theta \ell_i(\theta; x_i, y_i) \, \nabla_\theta \ell_i(\theta; x_i, y_i)^\top,
\end{equation}

where $\ell_i$ is the per-sample loss evaluated at the observed data label $y_i$ \cite{kunstner2019}. When the model is misspecified or far from convergence, the EF matrix can distort local curvature and bias update directions \cite{kunstner2019}. Second, practitioners discard off-diagonal terms to enforce a purely diagonal representation, reducing computational cost to $\mathcal{O}(D)$. This step decouples interacting parameter directions, introducing diagonal truncation error.

\subsection{Adam as a Diagonal Empirical Fisher Optimizer}
The standard Adam optimizer tracks historical moving averages of the first and second moments of stochastic gradients \cite{kingma2014}, following the update rule

\begin{equation}
\theta_{t+1} = \theta_t - \alpha \frac{\hat{m}_t}{\sqrt{\hat{v}_t} + \epsilon},
\end{equation}

where $\hat{m}_t = m_t / (1 - \beta_1^t)$ and $\hat{v}_t = v_t / (1 - \beta_2^t)$ are built recursively from $m_t = \beta_1 m_{t-1} + (1-\beta_1)g_t$ and $v_t = \beta_2 v_{t-1} + (1-\beta_2)g_t^2$ \cite{kingma2014}.

The coordinate-wise denominator $\sqrt{\hat{v}_t} + \epsilon$ directly mirrors the square root of a diagonal empirical Fisher matrix \cite{kingma2014, hwang2024}. Hwang leveraged this parallel to model Adam as a natural gradient optimizer under a diagonal empirical Fisher metric, assuming zero momentum ($\beta_1 = 0$) \cite{hwang2024}. In practice, Adam deploys diagonal truncation, empirical substitution, and temporal lag simultaneously. This study isolates their combined geometric cost.

\subsection{The \(\gamma(\Delta\theta)\) Geometric Indicator}
To track how closely an update vector $\Delta\theta$ mirrors an exact natural gradient step, Wu et al. designed the scale-invariant geometric indicator \cite{wu2024}

\begin{equation}
\gamma(\Delta\theta) = \frac{(\Delta\theta^\top \mathbf{F} \Delta\theta)^{\frac{1}{2}}}{|\Delta\theta^\top \nabla_\theta \mathcal{L}(\theta)|}.
\label{eq:gamma}
\end{equation}

A value of $\gamma = 0$ indicates perfect directional alignment with the exact NGD path, whereas expanding values denote directional divergence \cite{wu2024}. Because it is scale-invariant, it isolates directional fidelity within the Riemannian geometry rather than capturing step-size magnitude \cite{wu2024}.

\section{Related Work}
Amari demonstrated that optimization along the true Riemannian metric yields invariant step properties \cite{amari1998}. To avoid the high scaling costs of the full Fisher matrix, early variants adopted diagonal approximations. However, Kunstner et al. proved that substituting observed labels for model-sampled expectations can compromise performance when the model is far from convergence or structurally misspecified \cite{kunstner2019}. Their analysis showed that the resulting empirical Fisher can deviate heavily from the true Hessian, systematically misleading the optimizer \cite{kunstner2019}.

To address these instabilities, Wu et al. identified the inversely-scaled projection issue, showing that empirical Fisher approximations over-index on well-fit samples with near-zero gradients \cite{wu2024}. They proposed the improved empirical Fisher (iEF) to restore stability via gradient rescaling, validating it using the $\gamma(\Delta\theta)$ indicator on full-matrix variants \cite{wu2024}. Separately, Hwang formalized the connection between Adam and natural gradients through the FAdam optimizer, proving that Adam tracks an idealized natural gradient update when momentum is deactivated \cite{hwang2024}. This paper bridges these lines of work. We apply the geometric framework of Wu et al. \cite{wu2024} to analyze Adam's complete update rule under active momentum, directly evaluating how its approximation errors interact.

\section{Method}
We deploy the evaluation framework of Wu et al. \cite{wu2024} to compute the directional alignment between Adam's parameter updates and the true natural gradient path. This analysis tracks the metric $\gamma(\Delta\theta)$ from Equation~\ref{eq:gamma}, evaluating the explicit optimizer updates side-by-side:

\begin{equation}
\Delta\theta_t^{\text{SGD}} = -\alpha \nabla_\theta \mathcal{L}(\theta_t),
\end{equation}
\begin{equation}
\Delta\theta_t^{\text{EF}} = -\alpha \left[ \text{diag}(\hat{\mathbf{F}}^{\text{EF}}_t) + \epsilon \mathbf{I} \right]^{-1} \nabla_\theta \mathcal{L}(\theta_t),
\end{equation}
\begin{equation}
\Delta\theta_t^{\text{iEF}} = -\alpha \left[ \text{diag}(\hat{\mathbf{F}}^{\text{iEF}}_t) + \epsilon \mathbf{I} \right]^{-1} \nabla_\theta \mathcal{L}(\theta_t),
\end{equation}
\begin{equation}
\Delta\theta_t^{\text{Adam}} = -\alpha \left[ \text{diag}(\hat{\mathbf{F}}^{\text{EF}}_t)_{\text{EMA}} + \epsilon \mathbf{I} \right]^{-1/2} \left[ \nabla_\theta \mathcal{L}(\theta_t) \right]_{\text{EMA}},
\end{equation}

where the subscript $\text{EMA}$ denotes historical exponential moving averages. 

We evaluate these updates across four loss landscapes designed to isolate specific curvature and noise attributes: well-conditioned linear regression, ill-conditioned linear regression, logistic classification, and a non-convex multi-layer neural network. Our primary objective is to test whether landscape complexity accelerates geometric drift and to determine if iEF adjustments maintain alignment when paired with active momentum.

\section{Experiments}
Our experiments follow a progressive sequence across four loss landscapes to isolate curvature asymmetry, label substitution noise, and non-convexity. We track directional geometric deviation $\gamma(\Delta\theta)$ and training loss. The results demonstrate that Adam's geometric drift scales sharply with landscape complexity, yet this high tracking error does not compromise final loss minimization, as Adam consistently reaches identical or competitive loss to exact NGD. Furthermore, the improved empirical Fisher (iEF) tracks more stable paths than the volatile standard empirical Fisher (EF). To isolate these properties cleanly, learning rates are selected to optimize baseline stability rather than absolute convergence speed.

\subsection{Learning Rate Summary}
\begin{table}[H]
\centering
\setlength{\tabcolsep}{5pt}
\caption{Learning rates optimized for baseline stability within each setting.}
\label{tab:learning_rates}
\begin{tabular}{lcccc}
\toprule
\textbf{Optimizer} & \textbf{Well-cond.} & \textbf{Ill-cond.} & \textbf{Logistic} & \textbf{Neural Net} \\
\midrule
SGD  & $0.01$    & $0.0001$ & $0.01$ & $0.001$ \\
Adam & $0.01$    & $0.005$  & $0.01$ & $0.01$ \\
EF   & $0.01$    & $0.005$  & $0.01$ & $0.01$ \\
iEF  & $0.01$    & $0.005$  & $0.01$ & $0.01$ \\
NGD  & $0.1$     & $0.1$    & $0.1$  & $0.05$ \\
\bottomrule
\end{tabular}%
\end{table}
NGD consistently tolerates larger learning rates due to full-matrix curvature correction, whereas SGD becomes unstable under severe ill-conditioning.

\subsection{Experiment 1; Well-Conditioned Linear Regression}
This baseline consists of a linear regression task with $D=10$ independent features drawn from $\mathcal{N}(0,1)$, establishing an isotropic loss landscape where the exact Fisher matrix remains close to the identity matrix.

\begin{figure}[H]
    \centering
    \includegraphics[width=\textwidth]{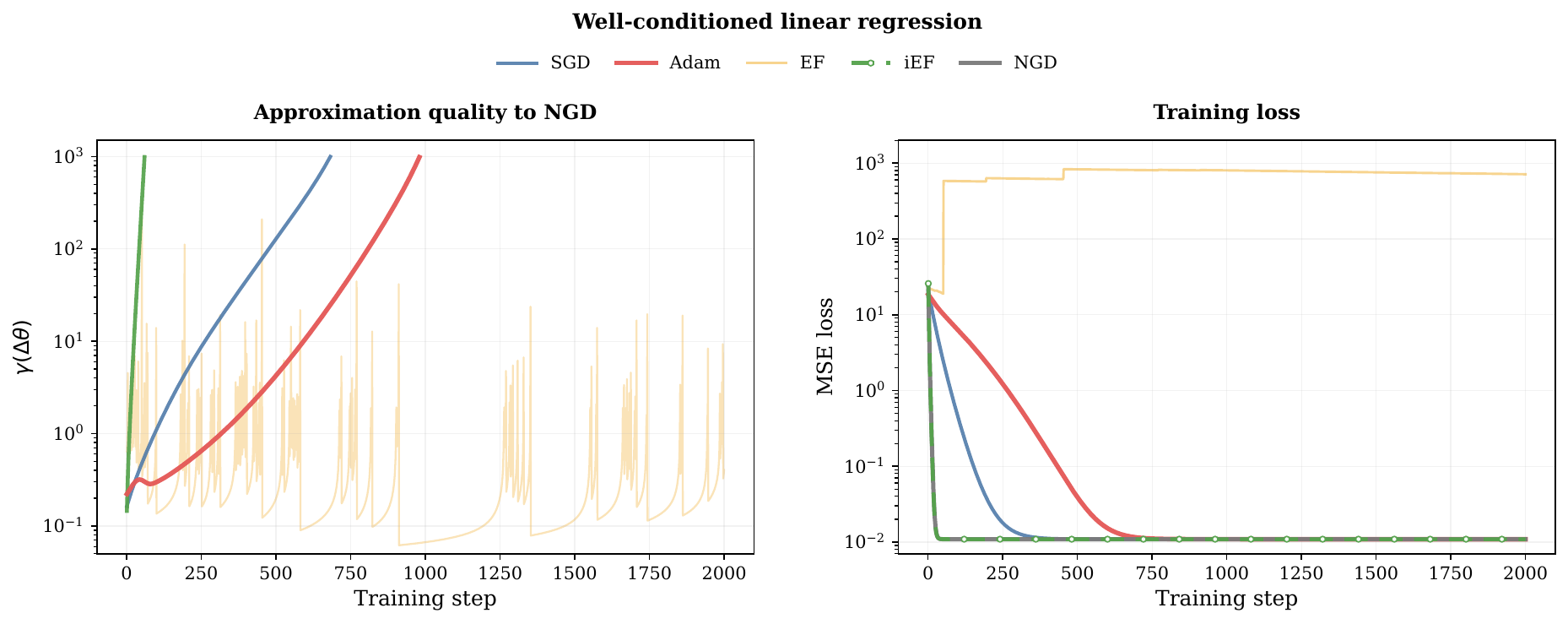}
    \caption{Geometric deviation $\gamma(\Delta\theta)$ (left) and training loss (right) for well-conditioned linear regression. Adam, SGD, and iEF exhibit rising geometric drift over 2000 steps but achieve lower loss floors than EF.}
    \label{fig:exp1}
\end{figure}

\subsection{Experiment 2; Ill-Conditioned Linear Regression}
To evaluate optimization under severe curvature heterogeneity, features are scaled to generate a $10{,}000\times$ variance asymmetry between parameter groups. This configuration creates a design covariance matrix $\Sigma = X^\top X$ with extreme eigenvalues. The resulting condition number is

\begin{equation}
\kappa(X^\top X) = \frac{\lambda_{\max}}{\lambda_{\min}} = \frac{15{,}237.35}{0.56} \approx 27{,}146.
\end{equation}

\begin{figure}[H]
    \centering
    \includegraphics[width=\textwidth]{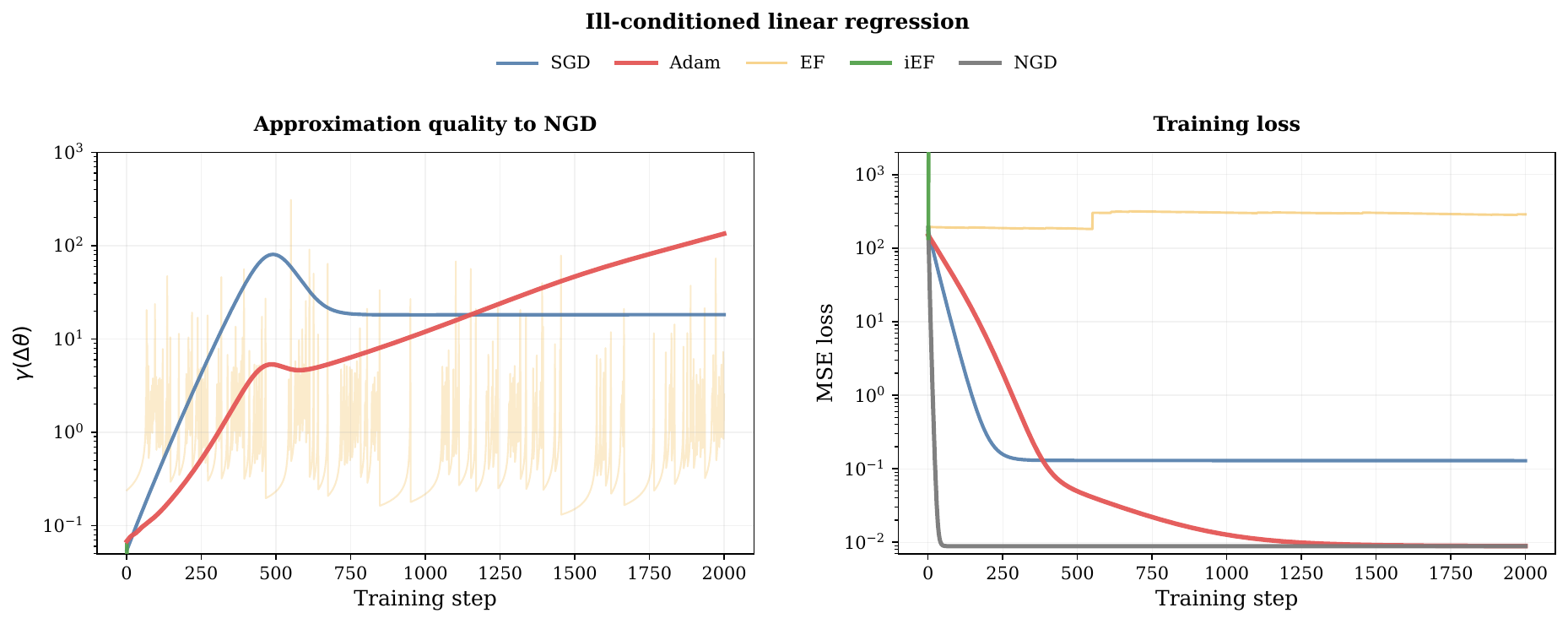}
    \caption{Geometric deviation $\gamma(\Delta\theta)$ (left) and loss curves (right) under severe ill-conditioning ($\kappa \approx 27{,}146$). Adam's geometric error rises steadily, crossing SGD near step 1250, while iEF maintains the lowest drift among the non-baseline approximations.}
    \label{fig:exp2}
\end{figure}

\subsection{Experiment 3; Logistic Regression}
This regime implements a binary classification task ($N=200, D=10$) using cross-entropy. It introduces structural model misspecification since the empirical Fisher evaluates curvature using observed data labels rather than model-predicted statistical expectations.

\begin{figure}[H]
    \centering
    \includegraphics[width=\textwidth]{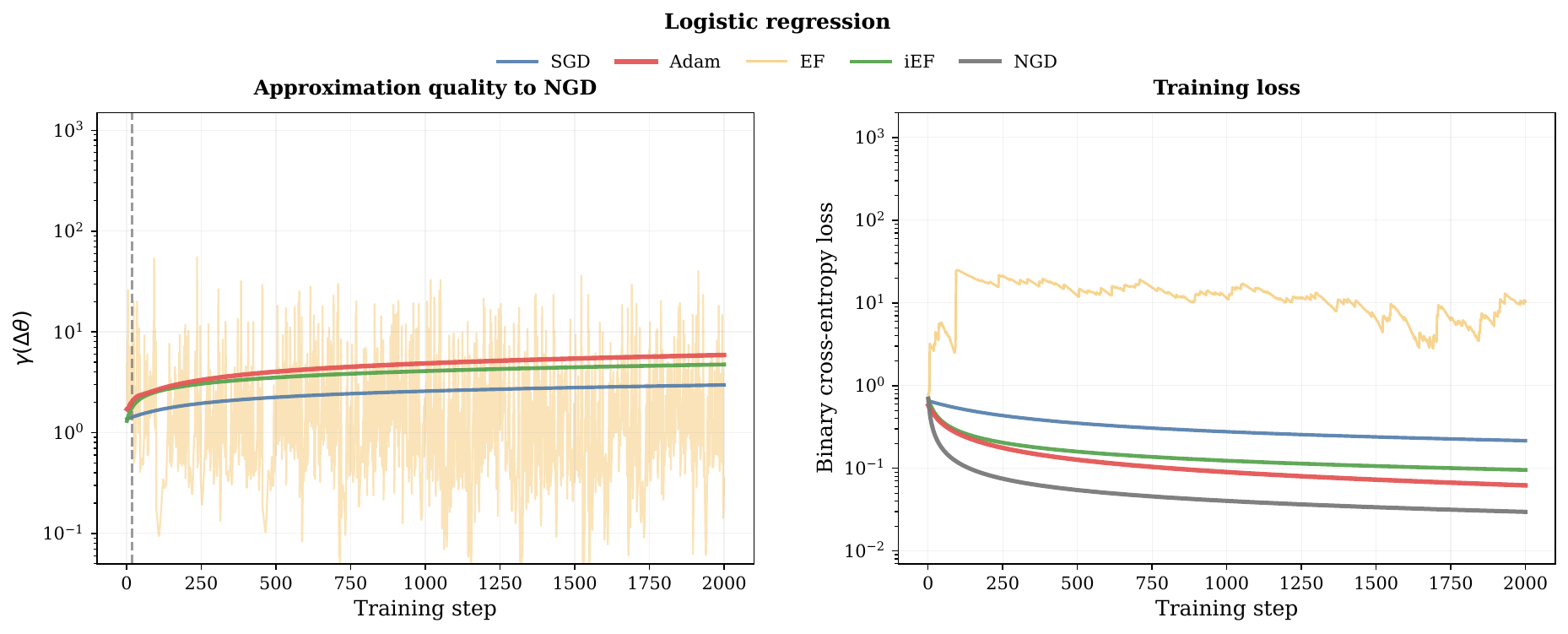}
    \caption{Metrics for logistic regression where label substitution distorts curvature. Adam exhibits higher initial geometric misalignment than SGD or iEF, while EF exhibits continuous oscillations.}
    \label{fig:exp3}
\end{figure}

\subsection{Experiment 4; Small Neural Network}
This non-convex regime uses a two-layer ReLU multilayer perceptron ($D_{\text{in}}=20, D_{\text{hidden}}=32, D_{\text{out}=1}$), yielding a 692-parameter space optimized against a fixed target network. It subjects the optimizers to all approximation errors simultaneously.

\begin{figure}[H]
    \centering
    \includegraphics[width=\textwidth]{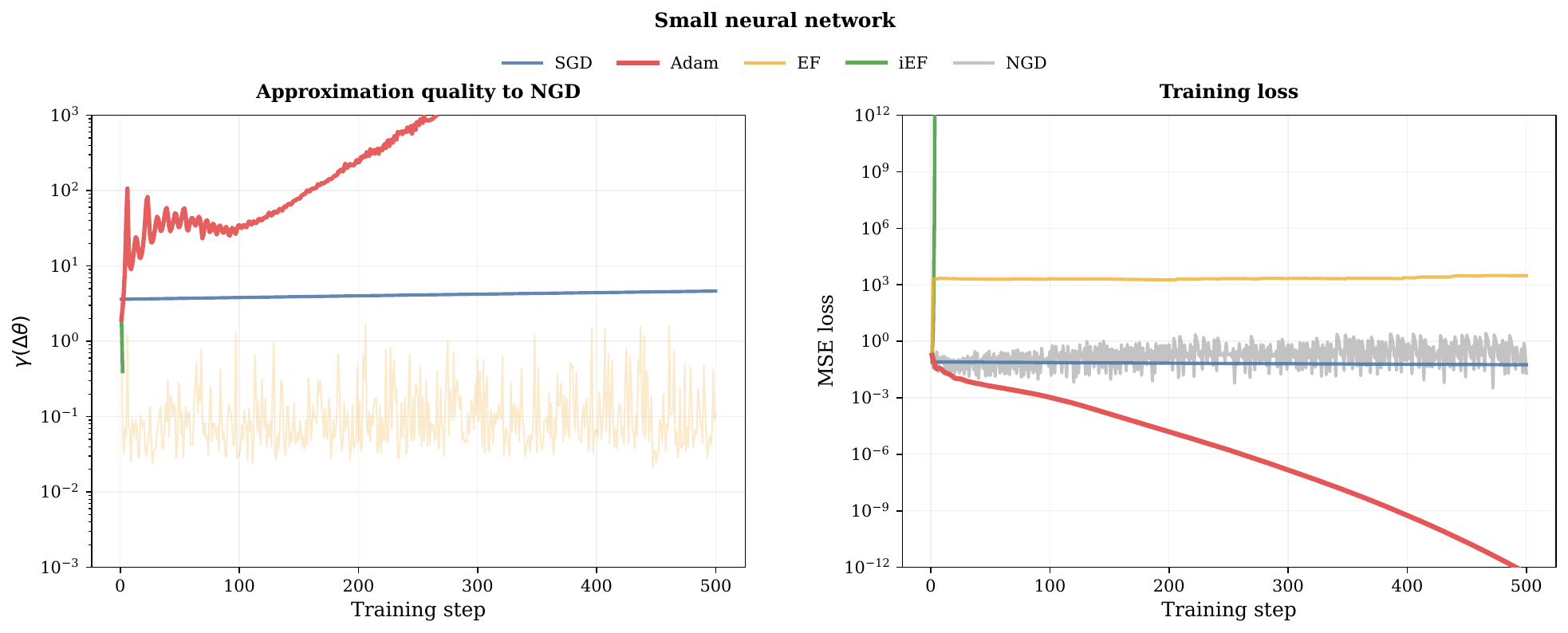}
    \caption{Optimization profiles within the non-convex neural network. Adam's geometric drift peaks near $\gamma \approx 10^3$ at step 300, yet it secures the lowest final training loss floor.}
    \label{fig:fig4}
\end{figure}

\section{Results}
Table~\ref{tab:summary} details the quantitative tracking data gathered across the four loss landscapes. The trajectories show that Adam's geometric drift $\gamma(\Delta\theta)$ increases alongside landscape complexity.

\begin{table}[H]
\centering
\caption{Quantitative tracking metrics and convergence thresholds across the evaluated environments.}
\label{tab:summary}
\resizebox{\textwidth}{!}{%
\begin{tabular}{lcccc}
\hline
\textbf{Experiment} & \textbf{Adam $\gamma$ (step 1)} & \textbf{Adam $\gamma$ (final)} & \textbf{NGD conv. step} & \textbf{Adam conv. step} \\
\hline
Well-conditioned linear & $\approx 0.2$ & $\approx 10^3$ & 18 & $\approx 600$ \\
Ill-conditioned linear  & $\approx 0.08$ & $\approx 10^2$ & 44 & $\approx 1000$ \\
Logistic regression     & $\approx 1.5$ & $\approx 6$ & 17 & $>2000$ \\
Neural network          & $\approx 3$ & $\approx 10^3$ & $>500^*$ & $<500$ \\
\hline
\multicolumn{5}{l}{$^*$NGD experiences numerical oscillations within the non-convex regime; see Section 6.4.}
\end{tabular}%
}
\end{table}
\subsection{Geometric Drift Across Landscapes}
Adam's initial geometric alignment depends on the loss landscape structure. In the isotropic setting, $\gamma$ starts low at $\approx 0.2$. Under severe ill-conditioning, alignment degrades during intermediate training phases, indicating that a diagonal preconditioner cannot fully compensate for scaling differences across coupled parameter groups. In logistic regression, the indicator starts at an elevated baseline of $1.5$, showing that empirical label substitution introduces directional distortion from the first epoch.

\subsection{Convergence Under High Geometric Error}
The neural network experiment highlights a strong separation between geometric alignment with NGD and successful optimization output. Adam's directional error climbs to $\approx 10^3$ by step 300, marking the highest tracking error observed. Yet, it simultaneously secures the lowest final training loss. This divergence suggests that $\gamma$ tracks directional compliance with NGD coordinates rather than predicting absolute convergence efficiency. Adam's momentum smoothing appears to enable it to navigate highly misaligned parameter updates successfully.

\subsection{Ill-Conditioned Crossover}
In Experiment 2, Adam's geometric error $\gamma$ surpasses that of standard SGD near step 1250, meaning it operates with less geometric fidelity than an un-preconditioned gradient step despite converging faster. This behavior may arise because the running diagonal empirical Fisher average overestimates local curvature along minimized, high-variance feature dimensions near the valley floor. Because off-diagonal terms are omitted due to diagonal truncation, the resulting preconditioner over-attenuates updates along these directions, biasing the update vector away from the true NGD path. This crossover is absent in the isotropic experiment, supporting the idea that curvature heterogeneity degrades diagonal approximations during late-stage convergence.

\subsection{EF vs iEF Stability}
The behavior of the standard empirical Fisher changes significantly in the non-convex regime. In Experiment 4, the EF method tracks a lower absolute $\gamma$ than SGD right before inducing optimization divergence, indicating that directional alignment metrics do not automatically guarantee stability. The iEF alternative maintains superior stability across Experiments 1, 3, and 4, though it falls short of Adam's optimization speed. Moreover, iEF experiences instability under extreme ill-conditioning, which suggests that its scaling adjustments require additional momentum smoothing in uneven landscapes.

\section{Discussion}

\subsection{How Adam's Approximations Interact}
Our experiments show that Adam drifts from NGD across all settings, yet it routinely matches or exceeds its convergence quality. To understand this, we examine how Adam's approximation mechanisms interact. Diagonal truncation removes off-diagonal interactions, causing errors under ill-conditioning where off-diagonal terms map parameter coupling. When combined with empirical label substitution, this distortion compounds as label noise enters the curvature estimate.

Crucially, this noisy, truncated preconditioner is smoothed by an exponential moving average, introducing temporal lag. This lag may help stabilize noisy updates. The moving average balances the directional errors introduced by coordinate truncation and empirical noise. Consequently, our results suggest Adam’s success may not depend on tracking NGD closely; instead, it relies on a balanced system where temporal smoothing mitigates approximation errors.

\subsection{Adam and Sign Gradient Descent}
Our results suggest that Adam's optimization mechanics may be more accurately modeled as a variant of sign gradient descent than as an approximation of NGD. SignGD updates using only gradient signs and discards magnitude information \cite{balles2018}. Within Adam's update loop, whenever the second-moment hyperparameter $\beta_2$ approaches unity, the coordinate-wise denominator $\sqrt{\hat{v}_t}$ stops tracking instantaneous local curvature and instead acts as a stable coordinate normalizer, while the first-moment average handles directional smoothing.

Our neural network evaluations match this interpretation; Adam maintains high geometric misalignment of $\gamma \approx 10^3$ yet minimizes the objective function faster than all alternative approaches. These findings are consistent with the view that Adam may behave similarly to sign gradient methods under coordinate-wise magnitude normalization and momentum smoothing rather than strict compliance with Riemannian metrics.

\subsection{Trade-Offs of Exact NGD}
In clean linear landscapes, NGD exploits exact analytical curvature; similarly, in logistic frameworks, it leverages true model-derived curvature fields. This access allows NGD to adjust each parameter step to the local geometry, reaching low loss in few steps. The primary limitation remains the high computational cost; storing and inverting full matrices scales poorly to large neural network architectures. Adam bypasses these hardware limits by utilizing a highly scalable $\mathcal{O}(D)$ diagonal layout, sacrificing fine-grained geometric alignment to preserve scalability.

\subsection{Limitations of $\gamma$}
The empirical dynamics observed in the neural network experiment indicate that while the \(\gamma(\Delta\theta)\) indicator provides valuable directional diagnostics, it remains an incomplete metric for overall optimizer health. The standard empirical Fisher regularly records lower absolute $\gamma$ values than standard SGD right before experiencing optimization divergence, demonstrating that tracking directional alignment cannot predict step-size stability or magnitude explosions. This represents a distinct boundary within the evaluation framework proposed by Wu et al. \cite{wu2024} that our experiments clarify. The metric captures directional geometry, but it lacks the mathematical machinery required to differentiate stable optimization steps from divergent ones.

\subsection{Practical Implications}
From a practical perspective, our findings indicate that the primary penalty associated with Adam's geometric approximation is an increased step count to reach convergence rather than any degradation in final loss quality. In three out of four evaluations, Adam eventually reaches the exact optimization floors secured by NGD, though it requires a longer training path. In the logistic regression setup, Adam fails to reach the baseline loss floor within a 2000-step window, indicating that model misspecification can limit final optimization performance under extended training budgets. For practitioners, these results reaffirm that Adam remains highly effective because it prioritizes computational efficiency and numerical stability.

\section{Future Work}
Our results suggest several future directions. First, integrating the geometric stability corrections of the iEF framework directly into Adam's momentum loop, an optimization variant we formally designate as IAdam, could capture both Riemannian geometric fidelity and practical convergence stability. Wu et al. explicitly deferred momentum integration to future studies \cite{wu2024}, and our results suggest that stabilized geometry-aware preconditioners may improve optimizer design. Second, extending these controlled geometric evaluations to overparameterized architectures and large image datasets, such as CIFAR-10, would determine whether the tracking patterns identified in our small-scale environments hold true across large-scale industrial deep learning workloads.

\section{Conclusion}
We applied the $\gamma(\Delta\theta)$ geometric tracking framework of Wu et al. \cite{wu2024} directly to Adam's complete update rule across four loss landscapes, quantifying the geometric costs of active momentum optimization under diagonal empirical Fisher approximations.

Our evaluations yield three primary conclusions. First, the $\gamma(\Delta\theta)$ metric is highly context-dependent, scaling up from a baseline of $\approx 0.2$ in well-conditioned linear setups to $\approx 10^3$ within non-convex neural networks as approximation errors overlap. Second, Adam regularly achieves stable convergence regardless of high geometric tracking error; its internal momentum provides numerical robustness that operates independently of strict geometric alignment. Our results are consistent with the view that Adam may behave similarly to sign gradient methods. Third, directional alignment metrics are informative but structurally incomplete; the standard empirical Fisher can achieve better directional alignment than SGD while simultaneously inducing optimization divergence, confirming that directional fidelity and step-size stability are fundamentally separate properties. In summary, our results suggest Adam’s practical optimization power may be enabled by its multi-layered approximations, trading strict theoretical geometry for the scalability that powers modern deep learning systems.

\section*{Acknowledgments}
The author thanks Antonio Orvieto for feedback on an earlier draft, including suggestions regarding sign gradient interpretations and the learning-rate summary table. Large language models were used for grammar checking, proofreading, and stress-testing arguments.

All source code required to reproduce the empirical curves presented throughout this paper is publicly available at the repository url \url{https://github.com/HobbesTheComputerScientist/adam-ngd-geometric-analysis}.

\end{document}